\documentclass[times, twoside, watermark]{zHenriquesLab-StyleBioRxiv}
\usepackage{blindtext}
\usepackage{lipsum}
\usepackage{caption}
\begin{document}

% insert title
\title{Mix, Align, Distil: Reliable Cross-Domain Atypical Mitosis Classification}
%\title{MAD: Mix, Align, Distil for Cross-Domain Atypical Mitosis Classification}
% insert title for footer
%\shorttitle{Approach for MIDOG 2022}

% Use letters for affiliations, numbers to show equal authorship (if applicable) and to indicate the corresponding author
\author[1]{Kaustubh Atey}
\author[2]{Sameer Anand Jha}
\author[2]{Gouranga Bala}
\author[1, 2]{Amit Sethi}

\affil[1]{Centre for Machine Intelligence and Data Science, IIT Bombay, India}
\affil[2]{Department of Electrical Engineering, IIT Bombay, India}

\maketitle

%TC:break Abstract
%the command above serves to have a word count for the abstract
% \begin{abstract}
% This is the abstract of your short paper describing your method contribution to the MIDOG 2022 MICCAI challenge. Make sure to summarize your contributions as well as your results in brief to help others and your reviewers to better understand your ideas and approaches. Please also make sure that you refer to the MIDOG 2022 challenge in title or abstract, so people know what this work is about.
% \lipsum[5]
% \end {abstract}

\begin{abstract}
Atypical mitotic figures (AMFs) are important histopathological markers yet remain challenging to identify consistently, particularly under domain shift stemming from scanner, stain, and acquisition differences. We present a simple training-time recipe for domain-robust AMF classification in MIDOG~2025 Task~2. The approach (i) increases feature diversity via style perturbations inserted at early and mid backbone stages, (ii) aligns attention-refined features across sites using weak domain labels (Scanner, Origin, Species, Tumor) through an auxiliary alignment loss, and (iii) stabilizes predictions by distilling from an exponential moving average (EMA) teacher with temperature-scaled KL divergence. On the organizer-run preliminary leaderboard for atypical mitosis classification, our submission attains balanced accuracy of $0.8762$, sensitivity of $0.8873$, specificity of $0.8651$, and ROC~AUC of $0.9499$. The method incurs negligible inference-time overhead, relies only on coarse domain metadata, and delivers strong, balanced performance, positioning it as a competitive submission for the MIDOG~2025 challenge.
\end {abstract}
%TC:break main
%the command above serves to have a word count for the abstract


\begin{corrauthor}
kaustubh.atey@iitb.ac.in
\end{corrauthor}

\section{Introduction}
Atypical mitotic figures (AMFs) reflect aberrant cell division with mis-segregation of genetic material and are associated with aggressive disease, higher metastatic risk, and poorer outcomes. Even expert pathologists find it difficult to consistently distinguish AMFs from normal mitoses, which motivates automated analysis. Beyond detection, reliable \emph{classification} of AMFs would enable reproducible assessment of tumor aggressiveness, yet prior work on this subtyping task remains limited. MIDOG~2025 Task~2 formalizes the problem as binary patch-level classification (atypical vs.\ normal mitosis) under realistic variability. Algorithms must address severe class imbalance, substantial intra-class morphological diversity, subtle inter-class differences, and pronounced domain shifts across tumors, species, stains, and scanners, while maintaining performance on unseen acquisition conditions. \\

The MIDOG~2022 report~\cite{aubreville2024domain} emphasized domain shift as a central obstacle: models trained on particular scanners, stains, laboratories, or species showed noticeable degradation on unseen settings. It recommended domain-aware experimental design, including patient or slide disjoint splits, domain-stratified validation, and containerized submissions, to achieve true cross-site robustness. Methodologically, successful entries relied on stain and color augmentation, stain normalization, and explicit domain generalization methods such as feature alignment or adversarial training, often combined with self-supervision, hard-negative mining, test-time augmentation, and ensembling. The report also noted calibration and threshold transfer issues across domains and highlighted annotation ambiguity for borderline cases, reinforcing the need for morphology-centric, domain-invariant representations and principled cross-domain evaluation. \\

We propose a lightweight training-time recipe for domain-robust atypical mitosis classification. The approach increases feature diversity via style perturbations in the backbone, aligns attention-refined features across sites using weak domain labels, and yields stable predictions by distilling from an EMA teacher. On the preliminary leaderboard, the model shows strong and balanced performance.

\section{Methodology}
\label{sec:methods}
To improve the generalization ability of our model across domains, we employ three components: (i) MixStyle layers inside the feature extractor to promote domain-agnostic, diverse feature extraction; (ii) a feature-alignment loss on backbone features to explicitly align representations from different domains; and (iii) knowledge distillation using an exponential moving average (EMA) teacher to obtain a model that remains stable across domains.

% 1. MixStyle
\subsection{Simulating Inter Domain Style Shifts}
\label{subsec:mixstyle}
We use \emph{MixStyle} \cite{zhou2021domain} to simulate inter-domain stain and contrast shifts directly in feature space while preserving class labels. In histopathology, cross-site variation largely stems from staining and imaging differences. By perturbing only the \emph{style statistics} (channelwise mean and variance) of intermediate features, MixStyle leaves morphology intact and pushes the encoder to rely on structure rather than colorimetric quirks. \\

\textbf{Formulation:} Given per-instance backbone features $x \in \mathbb{R}^{C \times H \times W}$ with channel-wise statistics $(\mu(x), \sigma(x))$, we draw $\lambda \sim \mathrm{Beta}(\alpha,\alpha)$ and form mixed statistics with another feature $x'$:
\[
\tilde{\mu}=\lambda\,\mu(x) + (1-\lambda)\,\mu(x') \qquad
\tilde{\sigma}=\lambda\,\sigma(x) + (1-\lambda)\,\sigma(x')
\]
and re-standardize
\[
\tilde{x}=\tilde{\sigma}\odot \frac{x-\mu(x)}{\sigma(x)+\varepsilon} + \tilde{\mu}.
\]
Here, $\varepsilon>0$ ensures numerical stability and $\odot$ is channelwise multiplication. Here $\tilde{x}$ is supervised with the source label, without any label blending.\\

By convexly mixing style statistics across instances, MixStyle broadens the training distribution along color and contrast axes that dominate cross-site variability while holding content fixed. This acts as an implicit domain generalizer, improving robustness on unseen centers and scanners with negligible overhead.

% 2. Feature Alignment Loss
\subsection{Feature Alignment with CBAM}
\label{subsec:feat-align}
To complement the primary binary cross-entropy (BCE) classification objective, we introduce an auxiliary \emph{feature alignment} loss that reduces inter-domain mismatch. Let $F \in \mathbb{R}^{B \times C \times H \times W}$ be encoder features extracted \emph{before} global average pooling (GAP). We refine them using the Convolutional Block Attention Module (CBAM)~\cite{woo2018cbam}:
\[
\hat{F} \;=\; \mathrm{CBAM}(F)
\]
which applies channel followed by spatial attention to emphasize morphology while suppressing stain and illumination artifacts. We impose alignment on $\hat{F}$ (instead of raw $F$) because attention-refined representations are easier to align across domains. These refined features are then passed to the classifier head. \\

\textbf{Domain-aware alignment.}
Using domain labels within each mini-batch, we reduce feature-distribution discrepancies across domains. Let $\hat{F}_i \in \mathbb{R}^{C\times H\times W}$ be the CBAM feature map for sample $i$ ($C$ channels, spatial size $H{\times}W$), and let $\mathcal{I}_d$ denote the index set of samples from domain $d\in\{1,\dots,D\}$. For channel $c$, define the per-sample channel descriptor by global average pooling (GAP)
\[
f_{i,c} \;:=\; \operatorname{GAP}\!\big(F_{i,c,:,:}\big)
\;=\;
\frac{1}{HW}\sum_{h=1}^{H}\sum_{w=1}^{W} F_{i,c,h,w}
\]
We then define the per-domain channel mean $\mu_{d,c}$ and the inter-domain mean $\bar{\mu}_{c}$ as follows:
\[
\mu_{d,c}=\frac{1}{|\mathcal{I}_d|}\sum_{i\in\mathcal{I}_d} f_{i,c}
\qquad\text{and}\qquad
\bar{\mu}_{c}=\frac{1}{D}\sum_{d=1}^{D}\mu_{d,c}
\]
Using these means, we compute the inter-domain variance as:
\[
s_c^{2}=\frac{1}{D}\sum_{d=1}^{D}\bigl(\mu_{d,c}-\bar{\mu}_{c}\bigr)^{2}
\]
We use a log-stabilized channel average as the alignment loss:
\[
\mathcal{L}_{\mathrm{align}}=\frac{1}{C}\sum_{c=1}^{C}\log\!\bigl(1+s_c^{2}\bigr)
\]
which %down-weights outlier channels while% 
penalizes large inter-domain discrepancies. At test time, the encoder followed by CBAM produces a refined, domain-invariant representation that is passed to the classifier head to make predictions.

% 3. KNowledge Distillation
\subsection{Knowledge Distillation with an EMA Teacher}
\label{subsec:kd}
To promote prediction stability across domains, we employ knowledge distillation (KD)~\cite{hinton2015distilling} with a teacher maintained as an exponential moving average (EMA) of the student~\cite{tarvainen2017mean}. After each student update $\theta_S^{(k)}$, the teacher is updated as
\[
\theta_T^{(k)} \;=\; m\,\theta_T^{(k-1)} \;+\; (1-m)\,\theta_S^{(k)} \quad m\in[0,1)
\]
Given student logits $z_S$ and teacher logits $z_T$, we form temperature-softened distributions
\[
p_S^{(T)} \;=\; \mathrm{softmax}\!\left(\frac{z_S}{T}\right)
\qquad
p_T^{(T)} \;=\; \mathrm{softmax}\!\left(\frac{z_T}{T}\right)
\]
and minimize the temperature-scaled KL divergence
\[
\mathcal{L}_{\mathrm{KD}} \;=\; T^2 \, \mathrm{KL}\!\big( p_T^{(T)} \,\|\, p_S^{(T)} \big)
\]
with gradients flowing only through the student. This EMA teacher acts as a temporal ensemble that smooths spurious domain-specific variations, encouraging the student to match stable targets rather than noisy, site-dependent signals. In practice, this reduces sensitivity to stain and illumination shifts while improving cross-site generalization. \\

\textbf{Training objective:} Our final objective combines the primary binary classification loss with the auxiliary alignment and distillation terms:
\[
\mathcal{L}_{\mathrm{total}}
\;=\;
\mathcal{L}_{\mathrm{cls}}
\;+\;
\lambda_{\mathrm{align}}\,\mathcal{L}_{\mathrm{align}}
\;+\;
\lambda_{\mathrm{KD}}\,\mathcal{L}_{\mathrm{KD}}
\]
Here, $\mathcal{L}_{\mathrm{cls}}$ is the BCE-with-logits loss on the classifier output, $\mathcal{L}_{\mathrm{align}}$ is the domain-aware alignment loss defined in Section~\ref{subsec:feat-align}, and $\mathcal{L}_{\mathrm{KD}}$ is the temperature-scaled distillation loss from the EMA teacher (Section~\ref{subsec:kd}). The weights $\lambda_{\mathrm{align}}, \lambda_{\mathrm{KD}} > 0$ balance the auxiliary terms.

\section{Implementation Details}
\label{sec:impl}
This section details on the datasets and splits used for training, validation, and testing, followed by the augmentation pipeline applied to all inputs. Next, we summarize the feature extractor configuration and its intermediate refinement prior to aggregation. Finally, we provide training details, including objectives, optimization settings and regularization.

% 1. Datasets
\subsection{Datasets and Splits}
\label{subsec:datasets}
We use the \textit{MIDOG 2025 Atypical Training Set}~\cite{weiss_2025_15188326} together with the \textit{MIDOG 2021 subset of AMi-Br}~\cite{bertram2025histologic} for model development. The combined pool is partitioned into training and validation sets with a 75/25 split, ensuring no data leakage between splits. For held-out evaluation, we use two unseen test domains: the \textit{TUPAC16 subset of AMi-Br} and the \textit{OMG-Octo Atypical} dataset~\cite{shen_2024_14246170}. These sets serve solely to assess generalization to unseen domains and are never used for tuning or selection. A model is considered for submission only if it improves on these held-out sets. \\

For domain supervision, domain labels are defined as follows: within MIDOG 2025 Atypical Training Set, each domain corresponds to a unique combination of \emph{Scanner}, \emph{Origin}, \emph{Species}, and \emph{Tumor}. The MIDOG 2021 subset of AMi-Br constitutes a separate domain. These domain labels are used exclusively for the domain alignment loss.

% 2. Data Augmentation
\subsection{Data Augmentation}
\label{subsec:augs}
To improve robustness while preserving mitotic morphology, we apply a compact augmentation suite covering staining, illumination, and mild geometric variability. \emph{Scale/geometry:} random resized cropping, horizontal/vertical flips, rotations, and mild perspective warp. \emph{Photometric:} color jitter, Gaussian blur, automatic contrast adjustment, and sharpness adjustment. \emph{Synthetic artifacts:} gamma adjustment, posterization, and solarization to mimic acquisition/compression effects. \emph{Coarse occlusion:} random erasing with multiple window sizes. \\

Collectively, these transforms broaden the training distribution to approximate cross-site shifts, promoting invariance to nuisance factors and improving generalization to unseen domains.

% 3. Training
\subsection{Model Architecture and Training}
\label{subsec:model-train}
\paragraph{Model architecture.}
We adopt a DenseNet-121 backbone~\cite{huang2017densely} for feature extraction, using its convolutional trunk to produce pre-pooling feature maps. To expose intermediate representations to style perturbations, we insert MixStyle blocks at early and mid stages (Beta mixing with $\alpha=0.1$). The resulting features are first refined with a CBAM module and then, after global average pooling, fed to the classification head.

\paragraph{Training setup.}
We optimize with AdamW (initial learning rate $10^{-3}$, weight decay $10^{-2}$), batch size $512$, and a \texttt{ReduceLROnPlateau} scheduler that lowers the learning rate when the validation metric stalls. Gradients are clipped at a max norm of $1.0$. A teacher network is maintained as an EMA of the student ($m{=}0.999$). Knowledge distillation uses temperature $T{=}2.0$ with base weight $0.5$, linearly warmed up over the first 10 epochs and applied every batch. Validation and early stopping monitor the EMA model’s metric. The feature-alignment weight $\lambda_{\mathrm{align}}$ follows the DANN annealing schedule \cite{ganin2016domain}, ramping from $0$ to $1$ across training.

\section{Results}
\label{sec:results}
\paragraph{Leaderboard performance (Track 2: Atypical Mitosis Classification).}
On the Preliminary Evaluation Phase leaderboard, our submission achieves the following organizer-reported metrics (see Table~\ref{tab:leaderboard}). These results indicate strong discrimination (ROC AUC = $0.9499$) with balanced operating characteristics (sensitivity = $0.8873$, specificity = $0.8651$), reflecting robust cross-domain generalization for atypical mitosis classification. \\

\begin{table}[ht]
\centering
\caption{Preliminary leaderboard results (Track 2)}
\label{tab:leaderboard}
\resizebox{\columnwidth}{!}{%
\begin{tabular}{lcccc}
\hline
Method & Balanced Acc. & Sensitivity & Specificity & ROC AUC \\
\hline
Baseline  & 0.7933        & 0.9014      & 0.6851      & 0.8859  \\
\hline
Ours  & 0.8762        & 0.8873      & 0.8651      & 0.9499  \\
\hline
\end{tabular}}
\end{table}

\section{Discussion}
\label{sec:discussion}
Our method addresses domain shift through three complementary components: (i) MixStyle, which perturbs feature-space style statistics to reduce sensitivity to stain and illumination variation; (ii) CBAM-refined feature alignment, which explicitly reconciles representations across domains; and (iii) EMA-teacher distillation, which supplies a temporally stable target without adding inference-time cost. \\

On the organizer-run preliminary leaderboard for atypical mitosis classification, our approach achieves strong and balanced results (Table~\ref{tab:leaderboard}), indicating robust generalization to unseen domains. These findings position our submission as a strong contender for the MIDOG 2025 challenge.

%\begin{acknowledgements}
%I thank the academy for this award.
%\end{acknowledgements}

\section*{References}
\bibliography{literature}

\end{document}